%% file: spconf.tex
\documentclass{article}
\usepackage{fancyhdr}
\usepackage{spconf,amsmath,array,graphicx}
\usepackage{tabularx}
\usepackage{makecell}
\usepackage{color}
\usepackage{multicol}
\usepackage{lipsum}
\usepackage{dirtytalk}
\usepackage[dvipsnames,table,table,xcdraw]{xcolor}
\usepackage{booktabs}
\usepackage{pgfplots}
\usepackage{multirow}
\usepackage{graphicx}
\usepackage{subcaption}
\usepackage{amsmath}

\include{math_commands}

\title{Neural Architecture Search with Multimodal Fusion Methods for Diagnosing Dementia}

\name{Michail Chatzianastasis$^{\star \dagger}$ \qquad Loukas Ilias$^{\star \ddagger}$ \qquad Dimitris Askounis$^{\ddagger}$ \qquad Michalis Vazirgiannis$^{\dagger}$}

\address{$^{\dagger}$DaSciM, LIX, École Polytechnique, Institut Polytechnique de Paris, France \\ $^{\ddagger}$DSS Laboratory, School of ECE, National Technical University of Athens, 15773 Athens, Greece}

\fancypagestyle{plain}{
 \fancyhf{}
 \fancyfoot[C]{\thepage}%
}
\fancypagestyle{firstpage}{
 \fancyhf{}
 \fancyfoot[C]{\small © 2023 IEEE. Personal use of this material is permitted. Permission from IEEE must be obtained for all other uses, in any current or future media, including reprinting/republishing this material for advertising or promotional purposes, creating new collective works, for resale or redistribution to servers or lists, or reuse of any copyrighted component of this work in other works.}%
}

\begin{document}
\ninept
\maketitle
\def\thefootnote{$\star$}\footnotetext{The first two authors contributed equally.}\def\thefootnote{\arabic{footnote}}

\begin{abstract}
Alzheimer's dementia (AD) affects memory, thinking, and language, deteriorating person's life.  
An early diagnosis is very important as it enables the person to receive medical help and ensure quality of life. 
Therefore, leveraging spontaneous speech in conjunction with machine learning methods for recognizing AD patients has emerged into a hot topic.
Most of the previous works employ Convolutional Neural Networks (CNNs), to process the input signal.
However, finding a CNN architecture is a time-consuming process and requires domain expertise.   
Moreover, the researchers introduce early and late fusion approaches for fusing different modalities or concatenate the representations of the different modalities during training, thus the inter-modal interactions are not captured. 
To tackle these limitations, first we exploit a Neural Architecture Search (NAS) method to automatically find a high performing CNN architecture.
Next, we exploit several fusion methods, including Multimodal Factorized Bilinear Pooling and Tucker Decomposition, to combine both speech and text modalities.
To the best of our knowledge, there is no prior work exploiting a NAS approach and these fusion methods in the task of dementia detection from spontaneous speech.  
We perform extensive experiments on the ADReSS Challenge dataset and show the effectiveness of our approach over state-of-the-art methods.
\end{abstract}
\section{Introduction}
\label{sec:intro}
\thispagestyle{firstpage}

Alzheimer's disease (AD) is a neurodegenerative disease and is the main cause of dementia. 
Dementia is progressive, thus there are three stages\footnote{https://www.alzheimers.org.uk/about-dementia/symptoms-and-diagnosis/how-dementia-progresses/progression-stages-dementia}, namely the early or mild stage, the middle or moderate stage, and the late or severe stage. 
Since no cure exists, the early diagnosis of dementia is of great clinical importance for stabilizing cognitive decline\footnote{https://www.dementia.org.au/information/diagnosing-dementia/early-diagnosis-of-dementia}.
In addition, since dementia affects the temporal characteristics of spontaneous speech \cite{10.3389/fnagi.2015.00195}, datasets have been developed \cite{luz20_interspeech}, where researchers employ deep learning methods, which can be used in clinical practice for supporting the diagnosis of dementia through spontaneous speech.

Several research works have been introduced, which employ Convolutional Neural Networks (CNNs) for classifying subjects into AD patients and non-AD ones. Specifically, some of them use as input to CNNs embeddings of transcript data, i.e., GloVE, word2vec, etc. \cite{10.3389/fcomp.2021.624558}. Other approaches use as input the raw audio signal \cite{cummins20_interspeech,10.3389/fpsyg.2020.623237}, while others transform the speech signal to log-Mel spectrograms and Mel-frequency Cepstral Coefficients (MFCCs) \cite{cummins20_interspeech,10.3389/fcomp.2021.624683,9383491}.
However, constructing high-performance deep learning models requires extensive engineering and domain knowledge. 
Neural Architecture Search (NAS) has emerged as class of approaches that automate the
generation of state-of-the-art neural network architectures, thus limiting the human effort \cite{elsken2019neural,liu2018progressive,Chatzianastasis_2021_ICCV}. A powerful NAS method, namely DARTS \cite{liu2018darts}, has achieved great discovered high-performance convolutional architectures for image classification problems. DARTS uses a continuous relaxation of the architecture representation and then applies gradient descent to discover the best architecture. In this work, we present the first study that incorporates DARTS into a neural network for recognizing dementia from spontaneous speech.

To integrate modalities of both speech and transcripts for detecting AD patients, several multimodal methods have been proposed.
However, these methods entail significant limitations, which need to be addressed. Specifically, the authors employ early \cite{martinc20_interspeech,pompili20_interspeech,10.3389/fnagi.2021.635945} or late fusion \cite{cummins20_interspeech,10.3389/fcomp.2021.624659,pappagari20_interspeech,sarawgi20_interspeech} approaches, or add/concatenate the vectors of the different modalities during training \cite{10.3389/fcomp.2021.624683}. Exploiting early fusion approaches means that the authors extract features from the different modalities and concatenate these features into one feature vector at the input-level. However, feature extraction constitutes a time-consuming process, since it requires domain knowledge. Employing late fusion approaches means that multiple models must be trained separately. Concatenating/adding the representation vectors of different modalities during training cannot fully capture the complex correlations between multimodal features. Thus, these approaches do not capture the inter-modal interactions. 

To address this limitation, we propose a multimodal neural network, where we pass each transcript through a BERT model \cite{devlin-etal-2019-bert} and obtain a text representation. Next, we convert audio files into images consisting of log-Mel spectrograms, delta, and delta-delta. We pass each image through the DARTS model. Finally, we exploit a variety of fusion methods for modelling the inter-modal interactions, including Tucker decomposition, a method based on the block-superdiagonal tensor decomposition, etc. To the best of our knowledge, this is the first study to propose such a framework, which combines a NAS approach, a language model, and fusion methods in an end-to-end neural network.

Our main contributions can be summarized as follows:
\begin{itemize}
\setlength{\itemsep}{0pt}
    \item We employ a neural architecture search approach, namely DARTS, to automatically generate the best CNN architecture. 
    \item We introduce several fusion methods for combining the representations of the CNN and the BERT model effectively.
    \item We perform extensive ablation studies to study the impact of the depth of the CNN architecture.
    \item We perform a series of experiments and show that our introduced architecture yields comparable performance to state-of-the-art approaches.
\end{itemize}

\section{Dataset}

The ADReSS Challenge Dataset \cite{luz20_interspeech} consists of a train and a test set, where the train set includes 78 AD and 78 non-AD patients, while the test set includes 24 AD and 24 non-AD ones. This dataset comprises speech recordings and their corresponding manual transcripts. Each participant describes the Cookie Theft picture from the Boston Diagnostic Aphasia Exam \cite{10.1001/archneur.1994.00540180063015}. It is worth noting that in contrast to other datasets, the ADReSS Challenge dataset minimizes the risks of biases in the prediction task. Specifically, it is matched for gender and age and contains equal number of AD and non-AD patients. Additionally, recordings have been acoustically enhanced with stationary noise removal and audio volume normalization has been also applied to control for variation. Last but not least, this dataset provides the opportunity for evaluating our methods in a complete subject-independent setting.


\section{Methodology}
\begin{figure*}[!htb]
     \centering
         \centering
         \includegraphics[width=\textwidth]{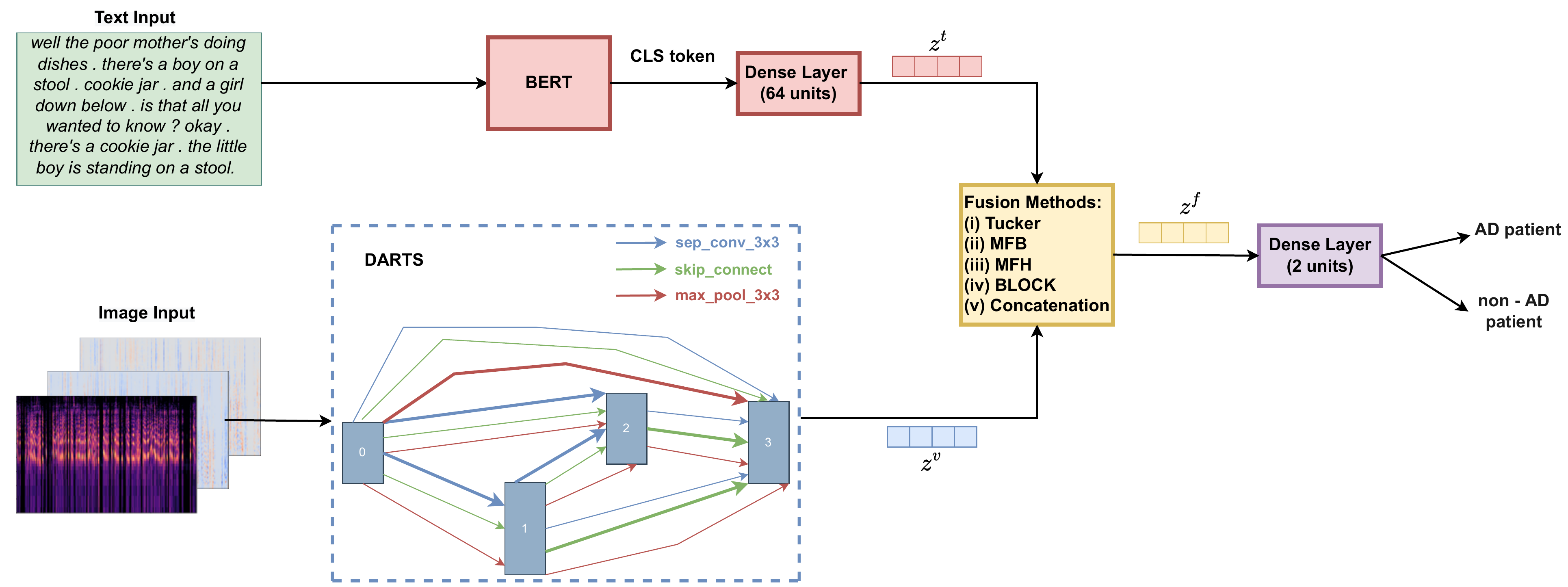}
         \caption{Illustration of our introduced architecture. For the text modality, we use a BERT language model to obtain the textual representation. In terms of the acoustic modality, we use the DARTS algorithm for obtaining the optimal CNN architecture and the acoustic representation. We fuse the two representations with fusion methods and pass the fused vector to a dense layer, which makes the prediction.}
         \label{fig:pipeline}
     \end{figure*}

In this section, we describe the functionality of the modules, which constitute our introduced architecture. 
First, we introduce the basic notation and describe the data preprocessing steps. Next, we present the neural architecture search algorithm, namely DARTS~\cite{liu2018darts} that automatically finds the best CNN architecture to process the input speech.
Then, we describe the module that process the text modality, using the BERT language model.
Finally, we present the multimodal fusion methods, that combine the two modalities and make the final prediction. 
The whole architecture is end-to-end trainable and is illustrated in Figure~\ref{fig:pipeline}. 

\noindent\textbf{Preliminaries.} Each input sample consists of a speech signal, a text description of the speech, and the label that indicates if the subject is an AD patient or a non-AD one.  We use \textit{librosa} \cite{mcfee2015librosa} and convert the audio files into images consisting of three channels, namely log-Mel spectrogram, delta, and delta-delta. We use 224 Mel bands, hop length accounting for 1024, and a Hanning window. Each image is resized to $224 \times 224$ pixels. We denote each image $i$ as $\bm{X_{I_i}} \in R^{224 \times 224 \times 3}$.

We exploit the python library called \textit{PyLangAcq} \cite{lee2016working} for reading the manual transcripts. We use the \textit{BertTokenizer} and pad each transcript to a maximum length of 512 tokens, while transcripts with number of tokens greater than 512 are truncated. \textit{BertTokenizer} returns the attention mask and the input\_ids per transcript. We denote each attention mask and input\_ids of a transcript $i$ as $\bm{X_{\alpha_i}} \in R^{512}$ and $\bm{X_{T_i}} \in R^{512}$ respectively. We further denote the binary label of each sample $i$ as $y_i \in \{0,1\}$.
Therefore each sample $i$ is represented as the tuple $\bm{(X_{I_i},X_{\alpha_i},X_{T_i}},y_i)$.
Our goal is to learn a function $f(\bm{X_{I_i},X_{\alpha_i},X_{T_i}})$, that takes as input the speech and text sample, and predicts the label of the subject.

\noindent\textbf{Speech-Neural Architecture Search.} CNNs have achieved great performance in image classification tasks, but require extensive architecture engineering. Therefore, in our work we aim to automatically learn the optimal CNN architecture using the DARTS model \cite{liu2018darts}. Following previous works \cite{liu2018darts,zoph2018learning}, since the final CNN architecture can have many layers, to reduce the computation complexity of the model, we search for a computational cell and then we stack this cell many times to construct the CNN architecture. 

Each cell can be represented as a directed acyclic graph (DAG), with 7 nodes. Every node $x_i$ denotes a feature map, and every edge $(i,j)$ transforms $x_i$ based on the operation of the edge $o_{(i,j)}$. Since the cell is a DAG, there exists a topological ordering of the nodes. Therefore, we can compute the feature map of each node, based on all the predecessors nodes, using the following equations:
\begin{equation}
    x_j = \sum_{i<j} o_{(i,j)}(x_i)
\end{equation}
The goal of the NAS algorithm then is to learn the operations on the edges. In our settings, we search operations in the following set $O$: \{3 $\times$ 3 and 5 $\times$ 5 separable convolutions, 3 $\times$ 3 and 5 $\times$ 5
dilated separable convolutions, 3 $\times$ 3 max pooling, 3 $\times$ 3 average pooling, identity, and zero, which indicates no connection\}.

However, gradient-based optimization is not directly applicable in a discrete search space. Therefore, we apply a continuous relaxation in the search space, by learning a set of weights $a$ for each edge operation. The discrete choice of each operation is transformed to a softmax over all operations: 
\begin{equation}
    \hat{o}_{(i,j)} = \sum_{o\in O} \dfrac{exp(a_{o_{(i,j)}})}{\sum_{\hat{o} \in O} exp(a_{\hat{o}_{(i,j)}})} o(x)  
\end{equation}
To obtain the final CNN architecture, we replace each operation   $\hat{o}_{(i,j)}$ with the operation with the largest weight $o_{i,j} = \operatorname*{argmax}_{o \in O} a_{o_{(i,j)}}$.

\noindent\textbf{Text-Language Models.} Bidirectional Encoder Representations from Transformers (BERT) is a multi-layer bidirectional Transformer encoder. It is trained on masked language modeling, where some percentage of the input tokens are masked at random aiming to predict those masked tokens  based on the context only. We pass to the BERT model the attention mask and the input\_ids denoted by $\bm{X_{\alpha_i}} \in R^{512}$ and $\bm{X_{T_i}} \in R^{512}$ respectively. We extract the classification token denoted by [CLS], where its dimensionality is equal to 768. Finally, we project its dimensionality to $d=64$.

\noindent \textbf{Multimodal Methods.} 
 Let $z^t \in \mathcal{R}^{64}$ denote the representation vector of the textual modality. Let $z^v \in \mathcal{R}^{64}$ denote the representation vector of the acoustic modality, extracted by the output of the CNN.
We fuse the two modalities, i.e., textual and acoustic, denoted by the vectors $z^t$ and $z^v$ by employing the following fusion methods:
\begin{itemize}
\setlength{\itemsep}{0pt}
    \item Tucker decomposition \cite{8237547}: a bilinear interaction where the tensor is expressed as a Tucker decomposition.
    \item Multimodal Factorized Bilinear pooling (MFB) \cite{8334194}: This approach enjoys the dual benefits of compact output features of  Multimodal Lowrank Bilinear (MLB) pooling and robust expressive capacity of  Multimodal Compact Bilinear (MCB) pooling.
    \item Multimodal Factorized High-order pooling (MFH)
    \cite{8334194}: The MFH approach is developed by cascading multiple
    MFB blocks.
    \item BLOCK \cite{Ben-younes_Cadene_Thome_Cord_2019}: Block Superdiagonal Fusion framework for multimodal representation based on the block-term tensor decomposition \cite{doi:10.1137/070690729}. It combines the strengths of the Candecomp/PARAFAC (CP) \cite{carroll1970analysis} and Tucker decompositions. 
    \item Concatenation: We concatenate $z^t$ and $z^v$, as $p = [z^t,z^v]$, where $p \in \mathcal{R}^{128}$. We pass $p$ through a dense layer consisting of 16 units with a ReLU activation function.
\end{itemize}

Finally, we obtain the fused vector, denoted by $z^f \in \mathcal{R}^{16}$, and we pass it through a dense layer with two units, which makes the final prediction.
We optimize the model using gradient descent by minimizing the cross-entropy loss.

\section{Experiments}

\noindent \textbf{Baselines.} We compare our approach with \textbf{(i)} unimodal approaches employing only the textual modality, i.e., BERT \cite{9769980}, \textbf{(ii)} unimodal approaches employing only the acoustic modality, i.e., DARTS, AT-LSTM (x-vector) \cite{9747167}, ECAPA-TDNN \cite{9414147}, SiameseNet \cite{cummins20_interspeech}, x-vectors\_SRE\cite{pompili20_interspeech}, Acoustic+Silence \cite{pappagari20_interspeech}, YAMNet \cite{10.3389/fcomp.2021.624683}, Majority vote (Acoustic) \cite{10.3389/fcomp.2021.624659}, Audio (Fusion) \cite{9459113}, DemCNN \cite{10.3389/fpsyg.2020.623237}, CNN-LSTM (MFCC) \cite{9383491}, and \textbf{(iii)} Multimodal approaches employing both the textual and acoustic modality, i.e., Audio + Text (Fusion)\cite{9459113}, Fusion Maj. (3-best) \cite{cummins20_interspeech}, Fusion of system \cite{pompili20_interspeech}, GFI, NUW, Duration, Character 4-grams, Suffixes, POS tag, UD \cite{martinc20_interspeech}, Acoustic \& Transcript \cite{pappagari20_interspeech}, Dual BERT (Concat/Joint, BERT large) \cite{10.3389/fcomp.2021.624683}, Majority vote (NLP + Acoustic) \cite{10.3389/fcomp.2021.624659}.

\noindent \textbf{Experimental Setup.} We minimize the cross-entropy loss function. We use a batch size of 8. We train the models on the ADReSS Challenge train set and report their performance on the test set. We divide the train set into a train and a validation set. We train the model for 50 epochs. We choose the epoch with the smallest validation loss and evaluate the performance of the model on the test set. We repeat the experiments five times and report the mean and standard deviation. We use Weights \& Biases \cite{wandb} for tuning the hyperparameters. Specifically, we perform a random search to optimize the following hyperparameters: number of CNN layers, learning rate for CNN, learning for alpha parameters of DARTS, learning rate for BERT, weight decay, fusion hidden dimension.  We use the BERT base uncased version provided via the Transformers library \cite{wolf-etal-2020-transformers}. All models are created using the PyTorch library and trained in a single NVIDIA RTX A6000 48GB GPU.

\begin{table*}[!h]
\tiny
\centering
\caption{Performance comparison among proposed models and state-of-the-art approaches on the ADReSS Challenge test set. Reported values are mean $\pm$ standard deviation. Results are averaged across five runs. Best results per evaluation metric are in bold.}
\begin{tabular}{lccccc}
\toprule
\multicolumn{1}{l}{}&\multicolumn{5}{c}{\textbf{Evaluation metrics}}\\
\cline{2-6} 
\multicolumn{1}{l}{\textbf{Architecture}}&\textbf{Precision}&\textbf{Recall}&\textbf{F1-score}&\textbf{Accuracy}&\textbf{Specificity}\\
\midrule
\multicolumn{6}{>{\columncolor[gray]{.8}}l}{\textbf{Unimodal state-of-the-art approaches (only transcripts)}} \\
\textit{BERT \cite{9769980}} & 87.19 $\pm$3.25 & 81.66 $\pm$5.00 & 86.73 $\pm$4.53 & 87.50 $\pm$4.37 & 93.33 $\pm$5.65\\
\midrule
\multicolumn{6}{>{\columncolor[gray]{.8}}l}{\textbf{Unimodal state-of-the-art approaches (only Speech)}} \\
\textit{DARTS} & 70.04 $\pm$3.84 & 89.99 $\pm$2.04 & 76.09 $\pm$0.87 & 72.92 $\pm$2.28 & 62.3 $\pm$7.05\\
\textit{AT-LSTM (x-vector) \makecell[l]{\cite{9747167}}} & 66.00 & 69.00 & 67.00 & 67.00 & -\\
\textit{ECAPA-TDNN \makecell[l]{\cite{9414147}}} & - & - & - & 66.70 & -\\
\textit{\makecell[l]{SiameseNet \cite{cummins20_interspeech}}} & - & - & 70.80 & 70.80 & -\\
\textit{x-vectors\_SRE \cite{pompili20_interspeech}} & 54.17 & 54.17 & 54.17 & 54.17 & 54.17\\
\textit{\makecell[l]{Acoustic+Silence \cite{pappagari20_interspeech}}} & 70.00 & 58.00 & 63.00 & 66.70 & 75.00\\
\textit{YAMNet \cite{10.3389/fcomp.2021.624683}} & 64.40$\pm$3.93 & 73.40$\pm$8.82 & 68.60$\pm$4.84 & 66.20$\pm$4.79 & 59.20$\pm$7.73\\
\textit{\makecell[l]{Majority vote (Acoustic) \cite{10.3389/fcomp.2021.624659}}} & - & - & - & 65.00 & -\\
\textit{Audio (Fusion) \cite{9459113}} & - & 83.33 & - & 81.25 & 79.17 \\
\textit{DemCNN \cite{10.3389/fpsyg.2020.623237}} & 62.50 & 62.50 & 62.50 & 62.50 & 62.50 \\
\textit{\makecell[l]{CNN-LSTM (MFCC) \cite{9383491}}} & 82.00 & 38.00 & 51.00 & 64.58 & 92.00 \\
\midrule
\multicolumn{6}{>{\columncolor[gray]{.8}}l}{\textbf{Multimodal state-of-the-art approaches (speech and transcripts)}} \\
\textit{Audio + Text (Fusion) \cite{9459113}} & - & 87.50 & - & 89.58 & 91.67\\
\textit{Fusion Maj. (3-best) \cite{cummins20_interspeech}} & - & - & 85.40 & 85.20 & -\\
\textit{Fusion of system \cite{pompili20_interspeech}} & 94.12 & 66.67 & 78.05 & 81.25 & \textbf{95.83}\\
\textit{\makecell[l]{GFI,NUW,Duration,Character 4-grams,Suffixes,POS tag,UD \cite{martinc20_interspeech}}} & - & - & - & 77.08 & -\\
\textit{Acoustic \& Transcript \cite{pappagari20_interspeech}} & 70.00 & 88.00 & 78.00 & 75.00 & 83.00\\
\textit{Dual BERT \cite{10.3389/fcomp.2021.624683}} & 83.04 $\pm$3.97 & 83.33 $\pm$5.89 & 82.92 $\pm$1.86 & 82.92 $\pm$1.56 & 82.50 $\pm$5.53 \\
\textit{Majority vote (NLP + Acoustic) \cite{10.3389/fcomp.2021.624659}} & - & - & - & 83.00 & -\\
\midrule
\multicolumn{6}{>{\columncolor[gray]{.8}}l}{\textbf{Our Proposed Architecture}} \\
\textit{DARTS+BERT+Tucker Decomposition} & 89.16 $\pm$ 3.96 & 85.00 $\pm$ 6.24 & 86.73 $\pm$ 1.57 & 87.08 $\pm$ 0.83 & 89.16 $\pm$ 5.00\\
\textit{DARTS+BERT+MFB} & 91.29 $\pm$0.34 & 88.29 $\pm$3.13 & 89.80 $\pm$1.76 & 89.58 $\pm$1.86 & 91.66 $\pm$1.26 \\
\textit{DARTS+BERT+MFH} & \textbf{94.46} $\pm$ 3.38 & 86.66 $\pm$ 3.11 & 88.31 $\pm$ 0.71 & 88.74 $\pm$ 1.02 & 94.16 $\pm$ 3.34 \\
\textit{DARTS+BERT+BLOCK} & 94.09 $\pm$2.61 & \textbf{91.66} $\pm$6.97 & \textbf{91.94} $\pm$1.98 & \textbf{92.08} $\pm$1.56 & 94.16 $\pm$3.33 \\
\textit{DARTS+BERT+Concatenation} & 86.68 $\pm$3.35 & 90.83 $\pm$1.66 & 88.65 $\pm$1.36 & 88.33 $\pm$1.66 & 85.83 $\pm$4.25 \\
\bottomrule
\end{tabular}
\label{compare3}
\end{table*}

\noindent \textbf{Evaluation Metrics.} We evaluate our proposed approaches using Precision, Recall, F1-score, Accuracy, and Specificity. The dementia class is considered the positive one.

\noindent \textbf{Results.}
The results are reported in Table~\ref{compare3}. 
Regarding our proposed multimodal models, we observe that DARTS + BERT + BLOCK is our best performing model reaching Accuracy and F1-score up to 92.08\% and 91.94\% respectively. It surpasses the introduced multimodal models in Recall by 0.83-6.66\%, in F1-score by 2.14-5.21\%, and in Accuracy by 2.50-5.00\%. DARTS+BERT+MFB constitutes our second best performing model achieving an Accuracy of 89.58\% and an F1-score of 89.80\%. It outperforms the introduced models, except for DARTS+BERT+BLOCK, in F1-score by 1.15-3.07\% and in Accuracy by 0.84-2.50\%. In addition, DARTS + BERT + MFH and DARTS + BERT + Concatenation yield almost equal Accuracy results, with DARTS + BERT + MFH surpassing DARTS + BERT + Concatenation in Accuracy by 0.41\%. On the contrary, DARTS + BERT + Concatenation outperforms DARTS + BERT + MFH in F1-score by a small margin of 0.34\%. We speculate that DARTS+BERT+MFB performs better than DARTS+BERT+MFH, since the MFH approach is developed by cascading multiple MFB blocks, thus is more complex for our limited dataset. In addition, we observe that the fusion method of Tucker decomposition yields the worst results reaching Accuracy and F1-score up to 87.08\% and 86.73\% respectively. 

Compared with unimodal approaches (employing only text), we observe that our introduced approaches, except for DARTS + BERT + Tucker Decomposition, outperform BERT \cite{9769980}. Specifically, DARTS + BERT + BLOCK improves the performance obtained by BERT \cite{9769980} in Precision by 6.90\%, in Recall by 10.00\%, in F1-score by 5.21\%, in Accuracy by 4.58\%, and in Specificity by 0.83\%. At the same time, we observe that the standard deviations over five runs are lower than BERT in all the evaluation metrics, except Recall. 

Compared with unimodal approaches (employing only speech), we observe that DARTS + BERT + BLOCK surpasses these approaches in Precision by 12.09-39.92\%, in Recall by 1.67-53.66\%, in F1-score by 15.85-40.94\%, in Accuracy by 10.83-37.91\%, and in Specificity by 2.16-39.99\%. We also compare our best performing model with DARTS and show that our best performing model outperforms DARTS in Precision by 24.05\%, in Recall by 1.67\%, in F1-score by 15.85\%, in Accuracy by 19.16\%, and in Specificity by 31.86\%. Next, we compare our approach, i.e., DARTS, with the existing research initiatives employing only speech. We observe that DARTS outperforms all the research works, except for Audio (Fusion) \cite{9459113}, in Accuracy by 2.12-18.75\%. DARTS attains a Recall score accounting for 89.99\% and outperforms the state-of-the-art approaches, including Audio (Fusion), in Recall by 6.66-51.99\%. DARTS outperforms also the existing research initiatives in terms of F1-score by 5.29-25.09\%.

In comparison with multimodal state-of-the-art approaches, we observe that our best performing model outperforms the existing research initiatives in Recall by 3.66-24.99\%, in F1-score by 6.54-13.94\%, and in Accuracy by 2.50-17.08\%. Although Fusion of system \cite{pompili20_interspeech} obtains a better Specificity score by our best performing model, our best performing model surpasses this approach in Recall, F1-score, and Accuracy. It is worth noting that Recall is a more important metric than Specificity, since high Specificity and low Recall means that AD patients are misdiagnosed as non-AD ones.

\begin{figure}[t]
     \centering
     \begin{subfigure}[t]{0.49\columnwidth}
         \centering
         \includegraphics[width=\linewidth]{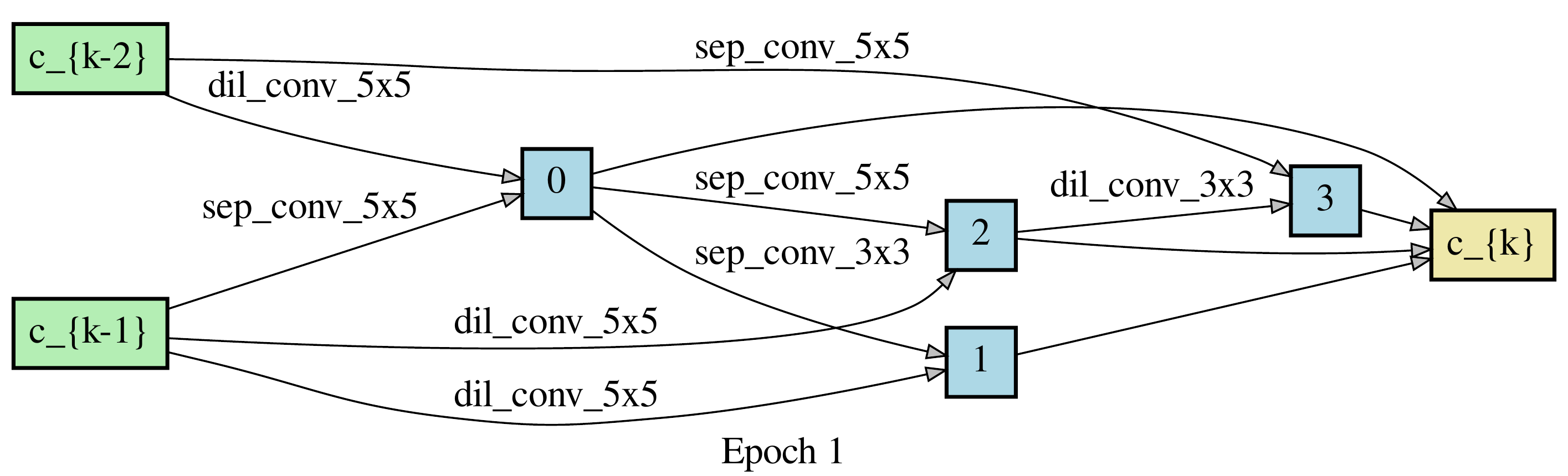}
         \caption{Normal Cell extracted from first epoch}
     \end{subfigure}
     \begin{subfigure}[t]{0.49\columnwidth}
         \centering
         \includegraphics[width=\linewidth]{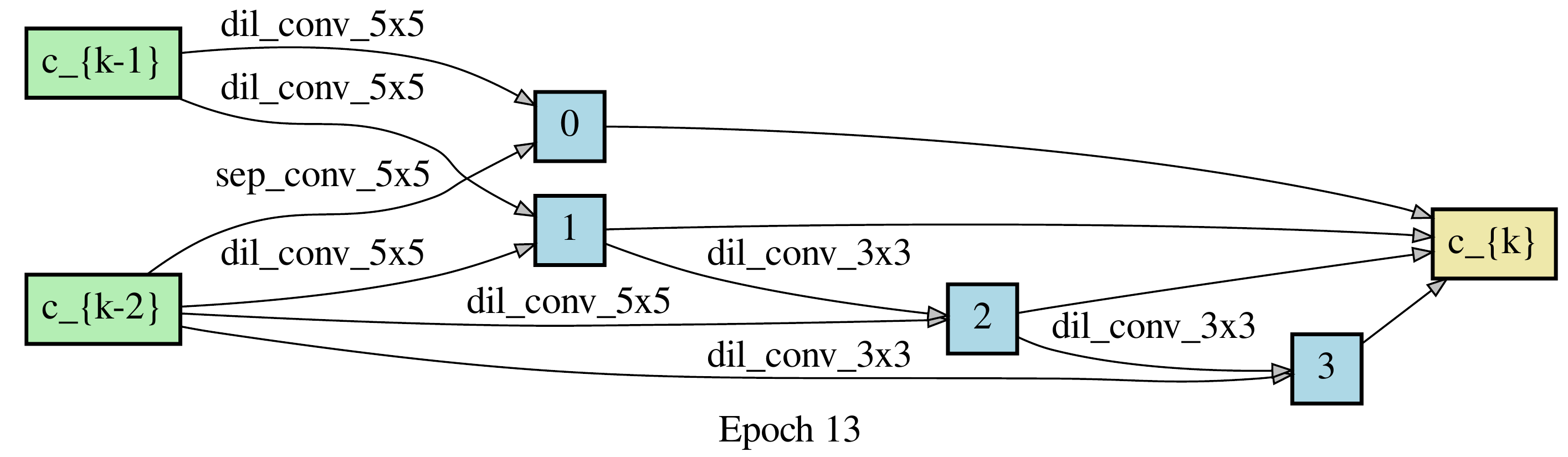}
         \caption{Best performing normal cell}
     \end{subfigure}
        \begin{subfigure}[t]{0.49\columnwidth}
         \centering
         \includegraphics[width=\linewidth]{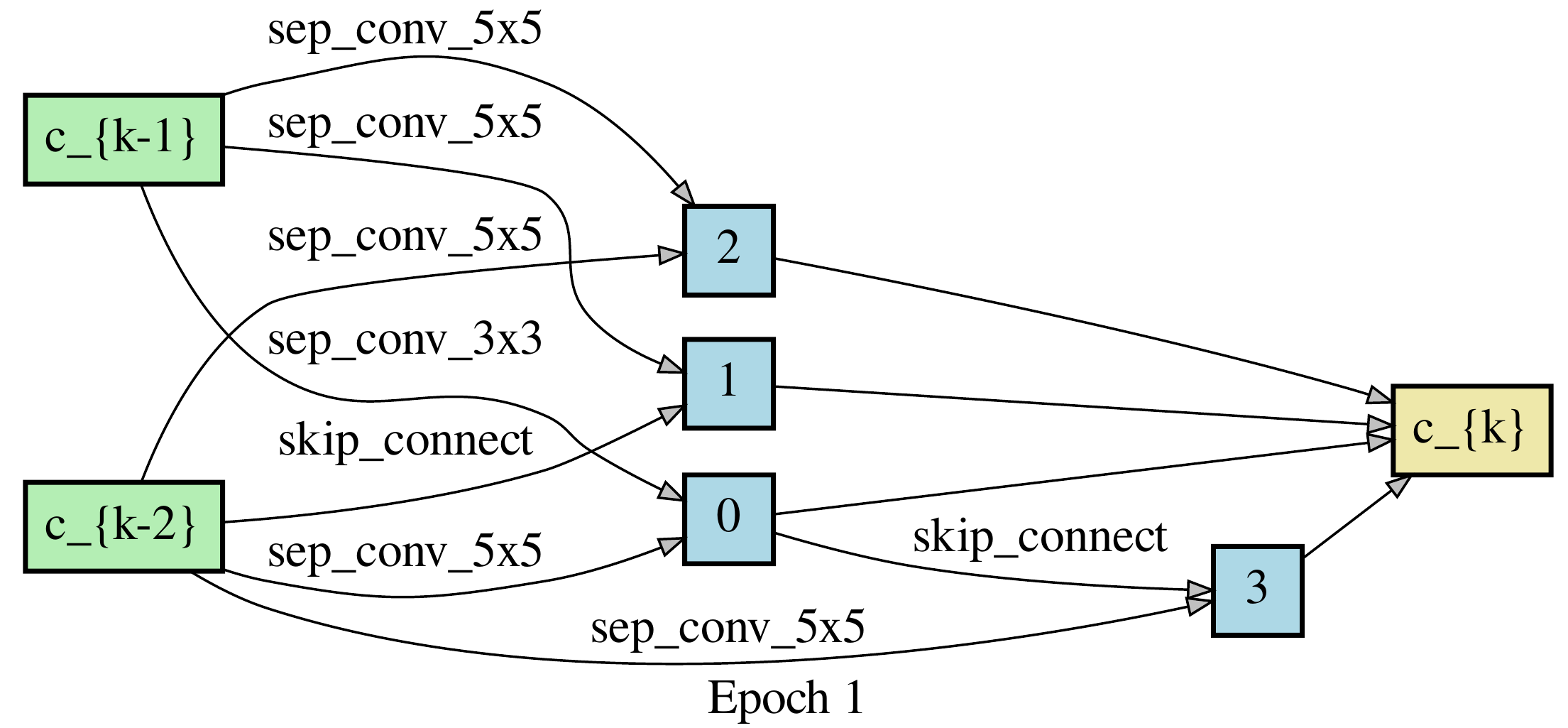}
         \caption{Reduce cell extracted from first epoch}
     \end{subfigure}
          \begin{subfigure}[t]{0.5\columnwidth}
         \centering
         \includegraphics[width=\linewidth]{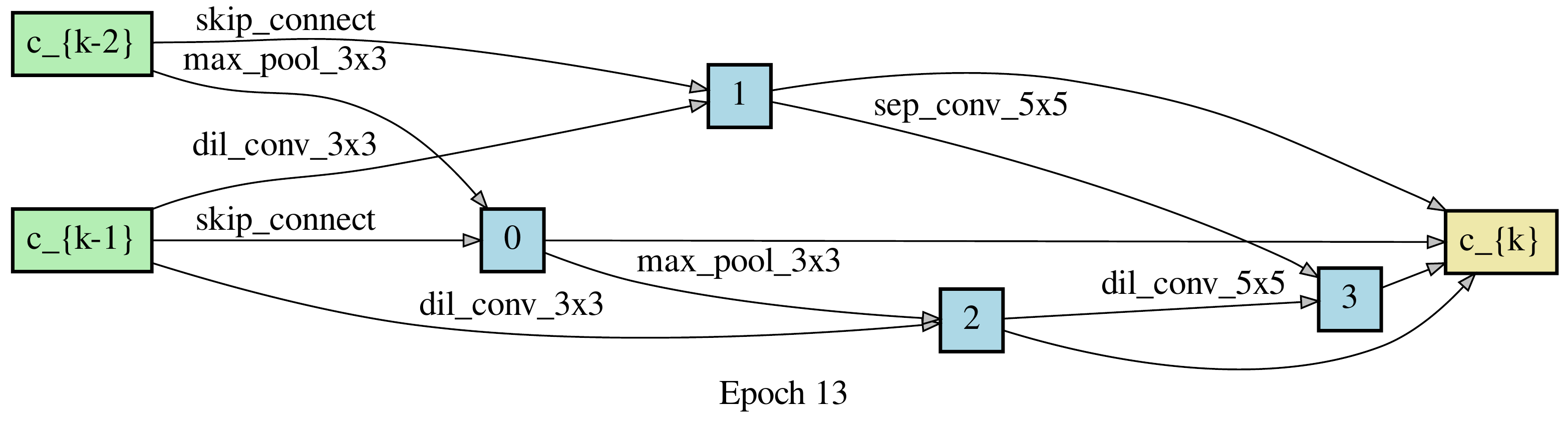}
         \caption{Best performing reduce cell}
     \end{subfigure}
     \caption{We visualize the initial normal and reduce cells and the best performing cells obtained from DARTS. These cells are stacked to create the convolutional neural network architecture.}
     \label{fig:best_arch}
\end{figure}

We further visualize the initialized and the best performing CNN architecture obtained by DARTS, in Figure \ref{fig:best_arch}. 
We observe that the best performing cell has different operations and different structure than the initial one, showing how the neural architecture search algorithm converges to an optimal cell, by altering the operations and the connections in the convolutional architecture.

\begin{figure}[t]
    \centering
    \includegraphics[width=0.3\textwidth]{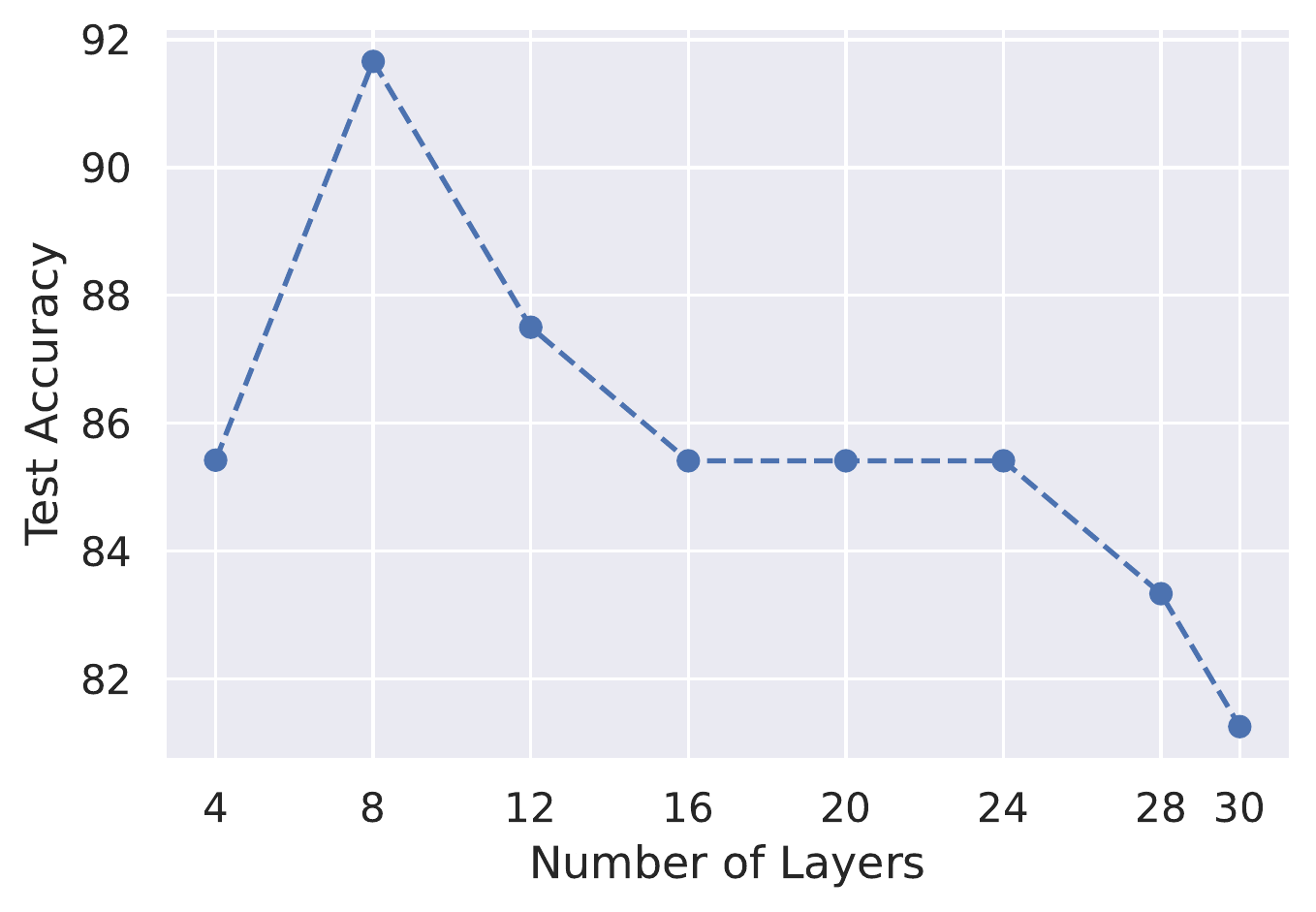}
    \caption{\label{fig:layers} Test accuracy of our proposed model with respect to the number of CNN layers generated from DARTS.}
\end{figure}

\noindent \textbf{Ablation Study.} We perform a series of ablation experiments, where we vary the layers of the CNN architecture, obtained by DARTS. Specifically, we set the number of CNN layers to 4, 8, 12, 16, 20, 24, 28, and 30. We report the accuracy obtained via these experiments in Fig.~\ref{fig:layers}. We observe that the best accuracy accounting for 91.66\% is obtained by using 8 layers. As the number of layers increases, the accuracy decreases. Specifically, the worst accuracy score is equal to 83.33\% and is obtained, when we use 30 CNN layers. 
We speculate that architectures with many layers are so complex for the dataset, and therefore the model overfits.

\vspace{-0.3cm}
\section{Conclusion and Future Work}

In this paper, we present the first study, which exploits Neural Architecture Search methods and fusion methods based on Tucker Decomposition, Factorized Bilinear Pooling, and block-term tensor decomposition, in the task of dementia detection. Specifically, we propose an end-to-end trainable multimodal model, which combines an automatically discovered CNN architecture obtained from the NAS algorithm as well as a language model for processing the text information. We integrate the two modalities using a variety of fusion methods. Our approach exhibits comparable performance with the state-of-the-art baselines. In the future, we plan to improve the NAS approach exploited in this study, by incorporating the fusion methods in the NAS pipeline. Also, we aim to apply explainability techniques, i.e., integrated gradients, GRAD-CAM, to explain the predictions of our proposed models. 

\bibliographystyle{IEEEbib}
\bibliography{strings,refs}

\end{document}

%% file: math_commands.tex
\usepackage{amsmath,amsfonts,bm}

\def\eqref#1{equation~\ref{#1}}

\def\1{\bm{1}}

\DeclareMathAlphabet{\mathsfit}{\encodingdefault}{\sfdefault}{m}{sl}
\SetMathAlphabet{\mathsfit}{bold}{\encodingdefault}{\sfdefault}{bx}{n}

